\documentclass{article}

\PassOptionsToPackage{numbers,sort&compress}{natbib}

\usepackage[main,final]{neurips_2026}

\usepackage[utf8]{inputenc} %
\usepackage[T1]{fontenc}    %
\usepackage{hyperref}       %
\usepackage{url}            %
\usepackage{booktabs}       %
\usepackage{amsfonts}       %
\usepackage{nicefrac}       %
\usepackage{microtype}      %
\usepackage{xcolor}         %

\usepackage{enumitem}
\usepackage{booktabs}
\usepackage{tabularx}
\usepackage{amsmath}
\usepackage{amssymb}
\usepackage{mathtools}
\usepackage{svg}
\usepackage{booktabs}
\usepackage{multirow}
\usepackage{makecell}
\usepackage{gensymb}
\usepackage{adjustbox}
\usepackage{wrapfig}
\usepackage[stable]{footmisc}

\usepackage{tikz}
\usetikzlibrary{decorations.pathreplacing,calligraphy,calc,positioning, tikzmark, intersections,fit}
\usepackage{xspace}

\usepackage{algorithm}
\usepackage{algpseudocode}
\usepackage{subcaption}

\newcommand{\Dim}[1]{\hfill $#1$}

\newcommand{\eqdef}{{\coloneqq}}

\newcommand\thsnd[2][000]{#2,#1}  

\newcommand{\sun}{S.U.N.\xspace}
\newcommand{\msun}{M.S.U.N.\xspace}
\newcommand{\morsun}{M/S.U.N.\xspace}

\title{SymADiT: Symmetry-Aware Crystal Generation with Latent Flow Matching}

\usepackage{authblk}

\author[1]{\textbf{Anmar Karmush}$^*$}
\author[2,3]{\textbf{Cedric Mathieu Brandenburg}}
\author[2,3]{\textbf{Soheil Ershadrad}}
\author[2,3]{\textbf{Johanna Rosen}} %
\author[1]{\textbf{Michael Felsberg}}
\author[4]{\textbf{Filip Ekström Kelvinius}}
\affil[1]{Department of Electrical Engineering (ISY) \& AI4x, Linköping University}
\affil[2]{Department of Physics, Chemistry and Biology (IFM), Linköping University}
\affil[3]{Wallenberg Initiative Materials Science for Sustainability (WISE), Linköping University}
\affil[4]{Department of Computer and Information Science (IDA), Linköping University}
\affil[*]{Corresponding author: \texttt{anmar.karmush@liu.se}}

\begin{document}

\maketitle

\begin{abstract} 

Tackling the task of crystal generation, we aim to enhance the previously proposed, symmetry-agnostic, All-atom Diffusion Transformer (ADiT) by introducing SymADiT, a symmetry-aware variant.
To do so, we use a representation of crystals based on Wyckoff positions. We follow ADiT and perform generative modelling in latent space, adapted to our symmetry-aware representation. 
By forcing the output of the generative model to adhere to the symmetry restrictions imposed by the generated crystal's space group and each atom's Wyckoff position, the generated crystals exhibit more realistic symmetry properties. 
We benchmark our method against both symmetry-aware and symmetry-agnostic models for crystal generation and show competitive performance, generating stable, symmetric, and physically plausible crystals with only an off-the-shelf Transformer architecture. Our code is available at \href{https://github.com/Karmush21/SymADiT}{https://github.com/Karmush21/SymADiT}.

\end{abstract}

\section{Introduction}
\label{sec:intro}
The vast combinatorial space of yet unexplored materials holds considerable potential for advancing several fields, including the discovery of novel 2D materials, next-generation energy storage systems, and materials for carbon capture \cite{intro_sentence}. In exploration of this space, generative models have emerged as an attractive class of methods \cite{gen_ai_review_paper}, largely due to their ability to rapidly generate large numbers of candidate materials. Within this broader materials space, crystalline materials are of particular interest, motivating substantial efforts toward their computational generation.

Early generative models for crystals \cite{cdvae, diffcsp,flowmm}, as well as more recent approaches \cite{adit, crystal_dit}, typically do not explicitly model the intrinsic symmetries of crystalline materials, instead relying on learning such symmetries directly from data. These symmetries are fundamental to determining material properties; for example, crystal symmetry governs whether phenomena such as piezoelectricity can occur \cite{example_of_property}. In addition, these \emph{symmetry-agnostic} methods exhibit two fundamental shortcomings: first, their representations are inherently redundant, as they include more structural detail than necessary, given that much of it is determined by the material’s symmetry; second, while these models are effective at generating novel crystal structures, the generated structures often exhibit only translational symmetry. This is highly implausible for real-world crystals and, from a generative modelling standpoint, inconsistent with crystal databases (e.g., \cite{cdvae, mpts52}), where usually only a small fraction of structures lack symmetry beyond pure translations.

To overcome these two issues, recent works have focused on \emph{symmetry-aware} generative modelling of crystals \cite{sgequi_diff, wyckoff_diff, symmcd, wyckoff_transformer, diffcsp++}. These methods adopt a \emph{Wyckoff}-based representation that explicitly encodes the crystal’s symmetry, retaining only the minimal information required to reconstruct the full structure via its symmetry operations. We contribute to this line of Wyckoff-based research by building on \emph{All-atom Diffusion Transformer} (ADiT) \cite{adit}, a recent generative framework for crystals and molecules. While ADiT employs a symmetry-agnostic representation of crystals, we propose \emph{SymADiT}, a symmetry-aware alternative that can be directly substituted into such a framework. In this work, however, we focus exclusively on  crystal generation. The central idea behind our representation is to restrict the model to predicting only the degrees of freedom (DOF) not constrained by the space group. By encoding symmetry directly through Wyckoff positions, the symmetry properties are enforced at the representation level rather than learned implicitly by the network. This allows the use of a standard off-the-shelf Transformer architecture \cite{transformer} while reducing the number of tokens relative to ADiT’s symmetry-agnostic representation.

\begin{figure}[t]
    \centering
    \includegraphics[width=\textwidth]{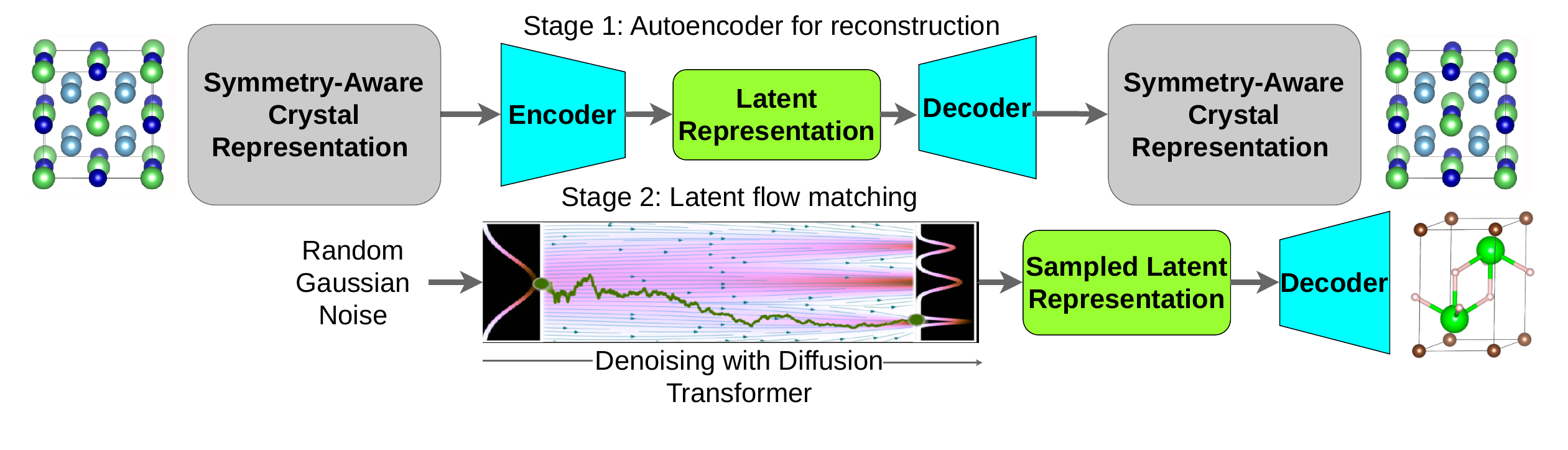}
        \vspace{-20pt} %
    \caption{Illustration of our representation and framework, which is inspired by ADiT \cite{adit}: Training an autoencoder for reconstruction, followed by learning a generative model over its latent representations. However, in our approach, the decoded representation explicitly contains symmetry information, thus enabling the generation of crystals with realistic symmetry properties by design.}
    \label{fig:proposed_framework}
    \vspace{-1.5em}
\end{figure}

Due to the nature of ADiT (which encodes both crystals and molecules), the generative modelling is performed in latent space. Following their design, we first train an autoencoder (AE) to encode crystals using our symmetry-aware representation, and subsequently perform generative modelling in the resulting latent space. This setting naturally enables us to investigate symmetry-aware generation in latent space, which, to the best of our knowledge, has not been previously explored. \footnote{Prior latent-space generative models have considered SE(3)-equivariance~\cite{GeoLDM},  in the context of molecular generation. Such models do not explicitly handle crystal-specific symmetries, such as space groups and Wyckoff positions. Here, \emph{symmetry-aware} refers specifically to the explicit modelling of the material symmetry properties, rather than only enforcing equivariance to rigid-body transformations.}
In contrast, most existing methods for crystal generation operate directly in data space. As a result, they typically require multiple generative components (e.g., separate mechanisms for continuous and discrete variables) \cite{sgequi_diff, MatterGen2025, cdvae}. By operating in latent space, the generative process is restricted to a single mechanism. The proposed methodology is illustrated in Figure \ref{fig:proposed_framework}.

In summary, our first contribution is a representation based on Wyckoff positions, which we incorporate into the ADiT framework to form \textit{SymADiT}, enabling symmetry-aware crystal generation with a simple Transformer backbone. We show how this enables more realistic generated crystals compared to ADiT, and we also benchmark our approach against a wide range of both symmetry-aware and symmetry-agnostic methods, showing competitive performance in crystal generation.

\section{Preliminaries: Wyckoff Representation}
\label{sec:preliminaries}
\paragraph{Symmetry-Agnostic Representation of Crystals}
An ideal crystalline material is represented by its crystal structure as an infinite periodic arrangement of \textit{unit cells}, allowing a single unit cell to serve as a finite representation. A widely used symmetry-agnostic approach \cite{cdvae, diffcsp, flowmm, adit, crystal_dit} is to represent the material by explicitly representing one full unit cell. In this way, a crystal $\mathbf{C}$ with $M$ atoms in its unit cell can be represented by a tuple $\mathbf{C} = (\mathbf{A}, \mathbf{X}, \mathbf{L})$, where $\mathbf{A} \in \mathbb{Z}^M$ denotes the chemical element associated with each atom, $\mathbf{X} \in \mathbb{R}^{M \times 3}$ are Cartesian atom coordinates, and $\mathbf{L} \in \mathbb{R}^{3 \times 3}$ are the unit cell basis vectors. The atomic positions can equivalently be expressed in the lattice basis as fractional coordinates $\mathbf{F} \in [0,1)^{M \times 3}$. In the remainder of this paper, we assume all atomic coordinates 
are given in fractional form. The lattice vector geometry can likewise be parametrised by six scalar parameters $\ell = (a, b, c, \alpha, \beta, \gamma)$, 
which provide a sufficient description\footnote{We lose the orientation of the cell with this parametrisation, but this can be arbitrarily chosen. This parametrisation of $\mathbf{L}$ is typically used in practice.} of $\mathbf{L}$. Since the choice of unit cell is not unique, prior symmetry-agnostic methods \cite{adit, cdvae, flowmm} typically canonicalize it using a primitive cell obtained via Niggli reduction \cite{niggli}. This symmetry-agnostic representation exhibits two main shortcomings (see Section \ref{sec:intro}), motivating the development of symmetry-aware approaches, which we discuss next and extend in our proposed representation.  
\paragraph{Space group}\label{par:space_group}
Fundamental for describing the symmetry of a crystal is its \emph{space group}. A space group $G \in \mathbb{G}$, where $\mathbb{G}$ denotes the set of $230$ space groups, is a group of isometries that tiles $\mathbb{R}^3$.  $G$ consists of two types of transformations: an infinite subgroup of discrete lattice translations, as well as a collection of symmetry operations $G(\cdot)=\{R_g(\cdot)+v_g | R_G \in O(3), v_g \in \mathbb{R}^3 \} \in G$, where $R_g$ is a point group operation (e.g., rotation, reflection), and $v_g$ is a translation. Once the space group is specified, much of the crystal’s structural information is largely constrained (see Paragraphs \nameref{par:wyckoff}, \nameref{par:unit_cell}, and Appendix \ref{app:example_of_rep}). This insight forms the basis of our representation, whose implications we discuss in the following sections.

\paragraph[Wyckoff Positions]{Wyckoff Positions\footnote{We cover only the essential aspects of Wyckoff positions needed to motivate our method; see, e.g., \cite{iuc, wyckoff_iuc_chapter} for a more comprehensive treatment.}}\label{par:wyckoff}
\definecolor{centeratom}{RGB}{60,91,140}
\definecolor{verticalatom}{RGB}{154,181,217}
\definecolor{generalatom}{RGB}{47,174,64}
\definecolor{diagonalatom}{RGB}{255,69,0}
\begin{wrapfigure}[25]{r}{0.4\textwidth} %
  \vspace{-0.5cm}
  \centering
    \begin{tikzpicture}[x=2cm,y=2cm]
    \draw[solid] (0,0) -- (2,0);
    \draw[solid] (0,0) -- (0,2);
    \draw[solid] (2,0) -- (2,2);
    \draw[solid] (0,2) -- (2,2);
    \draw[ultra thick, dashed, color=diagonalatom] (0,0) -- (1,1);
    \draw[ultra thick, dashed, color=verticalatom] (1,0) -- (1,1);
    \fill[gray!80,opacity=0.8] (0,0) -- (1,0) -- (1,1) -- cycle;

    \begin{scope}[
        atomnode1/.style={circle, draw=black, fill=centeratom, minimum size=15pt},
        atomnode2/.style={circle, draw=black, fill=diagonalatom, minimum size=13pt},
        atomnode3/.style={circle, draw=black, fill=generalatom, minimum size=10pt},
        atomnode4/.style={circle, draw=black, fill=verticalatom, minimum size=13pt},
    ]
    \node[atomnode1] at (1,1) {};
    \node[atomnode2] (leftorange) at (0,0) {};
    \node[atomnode3] at (0.5,0.2) {};
    \node[atomnode4] at (1,0.5) {};
    
    \end{scope}

    \begin{scope}[
        atomnode1/.style={circle, draw=black, fill=centeratom, minimum size=15pt, opacity=0.5},
        atomnode2/.style={circle, draw=black, fill=diagonalatom, minimum size=13pt, opacity=0.5},
        atomnode3/.style={circle, draw=black, fill=generalatom, minimum size=10pt,opacity=0.5},
        atomnode4/.style={circle, draw=black, fill=verticalatom, minimum size=13pt, opacity=0.5},
    ]

    \node[atomnode2] (rightorange) at (2-0,0) {};
    \node[atomnode3]  at (2-0.5,0.2) {};
    \node[atomnode3] at (2-0.2,0.5) {};
    \node[atomnode4] at (2-0.5, 1) {};

    \node[atomnode2] at (2-0,2-0) {};
    \node[atomnode3] at (2-0.5,2-0.2) {};
    \node[atomnode3] at (2-0.2, 2-0.5) {};
    \node[atomnode4] at (1,2-0.5) {};

    \node[atomnode2] at (0,2-0) {};
    \node[atomnode3] at (0.5,2-0.2) {};
    \node[atomnode3] at (0.2, 2-0.5) {};
    \node[atomnode4] at (0.5,1) {};

    \node[atomnode3] at (0.2, 0.5) {};

    \draw[->] (rightorange) to[bend left] node[above]{$M^{-1}$} 
    (leftorange);
    \draw[->] (leftorange) to[bend right=40] node[below]{$M$} (rightorange);

    \end{scope}
\end{tikzpicture}
\vspace{-0.30cm}
\caption{Toy illustration of Wyckoff positions in a 2D unit cell. Colours distinguish Wyckoff positions. Opaque atoms denote representatives, while faded atoms indicate symmetry-equivalent positions; together they form the corresponding orbit. M denotes a symmetry operation. The gray region highlights the asymmetric unit used in Wyckoff-based approaches, containing a single representative from each orbit. Image adapted from \cite{filip_thesis}.}
\label{fig:wyckoff_idea}
\end{wrapfigure}%

The Wyckoff positions of a space group $G$ partition regions of the unit cell with specific symmetry constraints.
When an atom occupies a Wyckoff position, the nature of that position constrains the coordinates to a point (0 DOF), a line (1 DOF), a plane (2 DOF), or a general position in the unit cell (3 DOF). More formally, the \emph{orbit} of a point $x$ is the set $\{gx | g \in G \}$, i.e., all positions obtained by applying every symmetry operation in $G$. All points in an orbit are symmetry-equivalent, and one element in that set is called a \textit{representative}. The number of distinct elements in the orbit is called the \textit{multiplicity}. Wyckoff positions are labelled by their multiplicity and a Wyckoff letter, with the letters assigned in lexicographic order corresponding (partially) to increasing multiplicity (e.g., \texttt{4a}, \texttt{8b}). These combinations are not unique across different space groups. A 2D illustration can be seen in Figure \ref{fig:wyckoff_idea}.
\paragraph{Unit Cell}\label{par:unit_cell}
 Similar to Wyckoff positions, the space group also constrains the unit cell geometry, meaning that certain parameters in $\mathbf{\ell}$ are no longer independent. For example, in cubic space groups (Nos.~195--230), the unit cell is constrained to be cubic, enforcing $a = b = c$ and $\alpha = \beta = \gamma = 90^\circ$, so that only a single lattice parameter remains independent ($a$ in $\ell$). In a Wyckoff-based representation, it is necessary to work with a larger symmetric unit cell, referred to as \emph{conventional}.
\paragraph{Asymmetric Unit} 
An alternative to the symmetry-agnostic representation (where the full unit cell is described) is to encode only a minimal subset of the unit cell, known as the \emph{asymmetric unit} (ASU), from which the full crystal can be reconstructed given the space group and its symmetry operations. This is the standard representation used in Wyckoff-based approaches. The ASU consists of one representative of each symmetry orbit
(see Figure \ref{fig:wyckoff_idea}). This shift to only considering symmetrically inequivalent atoms reduces the dimension of the generative modelling substantially. In Section \ref{sec:proposed_method}, we present our representation of the ASU, while prior approaches are reviewed in Section \ref{sec:related_works}.
\paragraph{Generative Models}
Generative frameworks \cite{diffusion_paper, discrete_diffusion_paper, score_based_paper, normalzing_flows, vae, gan} aim to learn the underlying (unknown) data distribution $p_\text{data}(\cdot)$ and generate new samples from it. In essence, the core idea is based on starting from a pure ``noise'' sample $x_0 \sim p_0(x_0)$, and turning this into  a ``clean'' 
sample $x_1 \sim p_\theta(x_1) \approx p_\text{data}(x_1)$, via a learned (deterministic or stochastic) map $x_1 = f_\theta(x_0)$.\footnote{Denoting noise as $x_0$ and the clean sample as $x_1$ is consistent with the framework we use, but the general description also covers other deep generative models.} The distribution $p_0(x_0)$ should be something we can easily sample, typically a standard Gaussian distribution. We use flow matching \cite{flow_matching} as our
generative framework, adjusted for operating in latent space, and describe this in more detail in Section \ref{subsec:ldm}.

\section{Proposed Method: SymADiT}
\label{sec:proposed_method}
In this section, we introduce our proposed method: Symmetry-aware ADiT (SymADiT). The method is based on a representation of crystals whose key idea is to exploit the fact that once the space group is specified, many structural degrees of freedom for the Wyckoff positions and unit cell become constrained, thereby reducing the dimensionality of the generative problem. Rather than predicting the full ASU structure, we instead model and predict only the symmetry-independent (\emph{free}) parameters, while all remaining quantities are deterministically inferred from the space group. Furthermore, by operating on symmetry-orbits rather than individual atoms, SymADiT substantially reduces the number of tokens processed by the Transformer compared with ADiT, resulting in a more compact representation and faster generation.

\begin{figure}[t]
\begin{minipage}[t]{0.48\textwidth}
\vspace{0pt}
\begin{algorithm}[H]
\scriptsize
\caption{Pseudocode for AE encoder $\mathcal{E}$}
\label{alg:ae_encoder}
\begin{algorithmic}[1]
\input{algorithms/encoder}
\end{algorithmic}
\end{algorithm}
\end{minipage}
\hfill
\begin{minipage}[t]{0.48\textwidth}
\vspace{0pt}
\begin{algorithm}[H]
\scriptsize
\caption{Pseudocode for AE decoder $\mathcal{D}$}
\label{alg:ae_decoder}
\begin{algorithmic}[1]
\input{algorithms/decoder}
\vspace{0.99cm} %
\end{algorithmic}
\end{algorithm}
\end{minipage}
\end{figure}
\subsection{Stage 1: Autoencoder}
A crystal structure\footnote{An example of how the representation is aligned with a crystal library (\texttt{PyXtal} \cite{pyxtal}) can be seen in Appendix \ref{app:example_of_rep}.}, $\mathbf{C}=(G, \mathbf{W,A,F}, \ell)$ consisting of $N$ symmetry orbits, can be represented via its ASU as:
\[
\begin{array}{c}
\text{Space group } G \in \{1,\dots,230\} \\[0.5em]
\begin{alignedat}{2}
\text{Atom types } \mathbf{A} &= \{a_i\}_{i=1}^N \in \mathbb{Z}^{N},
&\quad \text{Wyckoff pos. } \mathbf{W} &= \{w_i\}_{i=1}^N \in \mathbb{W}^{N}, \\[0.3em]
\end{alignedat}\\
\text{Fractional coords. } \mathbf{F} = \{f_i\}_{i=1}^{N} \in [0,1) ^{N \times 3}, \\ \quad
\text{Lattice params. } \ell = (a,b,c,\alpha,\beta,\gamma) \in \mathbb{R}^{6},
\end{array}
\]
where $\mathbb{W}$ denotes the set of all $1731$ Wyckoff positions across all space groups, and $\mathbb{Z}$ denotes the set of $100$ atom types considered in this work. Starting from the crystal ASU, an encoder $\mathcal{E}$ maps the attributes of each orbit to a latent representation $z_i$:
\begin{equation}
    \mathbf{Z} = \mathcal{E}(G, \mathbf{A},\mathbf{W},\mathbf{F},\ell), \ \ \ \ \ \ \ \ \ \mathbf{Z} = \{z_i\}^N \in \mathbb{R}^{N\times d},
\end{equation}
which is described in Algorithm \ref{alg:ae_encoder}.  Each latent vector of an orbit encodes both local information about the orbit $\mathbf{(A,W,F)}$, as well as global information about the crystal ($G,\ell$), with cross-orbit dependencies captured through self-attention \cite{transformer}. While ADiT uses KL divergence \cite{vae} as a regularizer to structure the latent space, we follow \cite{lost_in_latent_space} and replace this with a saturating function (adopting their setting, we fix the saturation parameter to $5$, see last line in Algorithm \ref{alg:ae_encoder}), which we found performed better. See Appendix \ref{app:ablation_studies} for an ablation study on different regularisation techniques. 

The decoder $\mathcal{D}$, described in Algorithm \ref{alg:ae_decoder}, maps the latent embedding back to components of $\mathbf{C}$ as
\begin{equation}
    \mathbf{A'}, \mathbf{F'}, \mathbf{W'}, \ell' = \mathcal{D}(\mathbf{Z}, G).
\end{equation}

Similar to ADiT~\cite{adit}, we use a standard off-the-shelf Transformer~\cite{transformer} for both the encoder $\mathcal{E}$ and decoder $\mathcal{D}$, with the only modification being the removal of unnecessary positional embeddings. The decoder predicts a full parameterization of the ASU, after which symmetry-constrained quantities are masked out or determined
deterministically from the space group and predicted Wyckoff positions to ensure adherence to symmetry. These operations are represented by the symmetrizing functions $\mathcal{S}_G\big[\cdot\big]$ and $\mathcal{S}_w\big[\cdot\big]$ in Algorithm~\ref{alg:ae_decoder}. For example, although the network predicts over the full enumeration of space group specific Wyckoff positions (1731 in total), only those compatible with the current space group are retained. Similarly, if a Wyckoff position permits only a single free coordinate $x$, the predicted $y$ and $z$ coordinates are excluded from decoding, with their values instead determined by the symmetry constraints. Likewise, for a cubic space group, only the lattice constant $a$ in $\ell$ constitutes an independent degree of
freedom; the remaining lengths $(b,c)$ and angles $(\alpha,\beta,\gamma)$ are determined by the space group constraints. Details on the network architectures are provided in Appendix~\ref{app:network_and_training}, while Appendix~\ref{app:latent_viz} visualizes the encoder's latent space via dimensionality reduction.

\paragraph{Training}
The autoencoder is trained using a weighted objective comprising a reconstruction loss and additional regularization terms.
The reconstruction loss itself consists of four terms, each corresponding to different components of the crystal's ASU:
\begin{align}
\mathcal{L} = \lambda_\mathbf{A}\mathcal{L}_\mathbf{A} + \lambda_\mathbf{W}\mathcal{L}_\mathbf{W} + \lambda_\mathbf{F}\mathcal{L}_\mathbf{F} + \lambda_\ell\mathcal{\ell}_\ell,
\quad (\lambda_A, \lambda_W, \lambda_F, \lambda_\ell) = (1,1,5,1).
\end{align}
As atoms and Wyckoff positions are discrete variables, their corresponding reconstruction terms $\mathcal{L}_\mathbf{A}$ and $\mathcal{L}_\mathbf{W}$ are standard cross-entropy losses. We use a periodic cosine loss for the fractional coordinates $\mathbf{F}$ and standard mean squared error (MSE) for the cell parameters $\ell$. The loss terms are weighted to emphasise accurate reconstruction of the fractional coordinates, with weights chosen empirically. Importantly, the loss is applied only to the free parameters of $\mathbf{W}$, $\mathbf{F}$, and $\ell$. Symmetry-constrained parameters are excluded from the loss, since their values are determined by the symmetry constraints and are not used during decoding.
\paragraph{Regularisation}
We utilise a number of soft regularisation techniques to improve the training of the autoencoder. Similar to ADiT \cite{adit}, we use a dimension of the latent representation $d$ that is significantly smaller than the encoder/decoder hidden dimension $d_m$.   
To reconstruct physically plausible crystals, the predicted ASU is expanded to the full crystal using the space group transformations, and the resulting structure is checked for interatomic distances below a specified threshold (0.5 Å). 
We apply augmentations to both the free lattice parameters and atomic positions to improve robustness. The perturbations are restricted to the unconstrained degrees of freedom imposed by symmetry. For example, in a cubic unit cell only the side-length ($a$ in $\ell$) is perturbed, while an atom constrained to move along a line (1 DOF) is perturbed only along that direction.

\subsection{Stage 2: Latent Flow Matching}
\label{subsec:ldm}
After the first training stage, the AE is frozen and its encoder $\mathcal{E}$ is used to produce latent representations of the crystals for training the latent flow matching model. The model learns to generate $N$ latent representations $\mathbf{Z} = \{z_i \in \mathbb{R}^d \}_{i=1}^N$, one per symmetry orbit in the ASU, where each $z_i$ encodes local and global structural information. Together, these can be decoded using the decoder $\mathcal{D}$ to produce a crystal's ASU. 

\paragraph{Training}
We closely follow the notation of ADiT \cite{adit}. During training, we begin by encoding a crystal's ASU into a latent representation $\mathbf{Z}$. We define $\mathbf{Z}^{(1)}$ as being a clean sample at time $t=1$. We then sample a noisy latent sample $\mathbf{Z}^{(0)} \sim \mathcal{N}(0,I_d)$ and construct an intermediate latent by linearly interpolating the two samples as
\begin{equation}
    \mathbf{Z}^{(t)} = (1-t)\mathbf{Z}^{(0)} + t\mathbf{Z}^{(1)}, \ \ t \in [0,1].
\end{equation}
This defines a continuous path from noise to data. Flow matching aims to learn a vector field $u_t(Z^{(t)})$ that describes how a sample should move along the path over time. The ground truth conditional vector field is given by 
\begin{equation*}
    u_t(\mathbf{Z}^{(t)} | \mathbf{Z}^{(1)}) = \frac{\mathbf{Z}^{(1)}-\mathbf{Z}^{(t)}}{1-t}, 
\end{equation*}
which corresponds to moving directly toward the clean latent sample. The goal of flow matching is to train a denoiser neural network $\mathcal{F}$ to match the conditional vector field. The network takes as input the intermediate sample, the time step, and additional information to predict a clean sample $\mathbf{Z}^{'(1)}$. Training of the denoiser network is done by minimizing MSE between $\mathbf{Z}^{(1)}$ and $\mathbf{Z}^{'(1)}$.

\paragraph{Generation}
During generation, we aim to sample from the (unknown) distribution $p_{\text{data}}(G, \mathbf{A}, \mathbf{W}, \mathbf{F}, \ell)$. Given that the space group constrains many degrees of freedom, we start by sampling a space group $G$. In practice, during crystal generation, each orbit will correspond to a token in the Transformer, and we therefore introduce the number of orbits $O$ as an auxiliary variable, which is sampled conditioned on the space group. We then sample the latents $\mathbf{Z}$ (which are also auxiliary variables) from our flow matching model, and the remaining variables are sampled by the (probabilistic) decoder $\mathcal{D}$ given $\mathbf{Z}$, $G$, and (implicitly, as it determines the number of orbits) $O$. For both $G$ and $O$, we can use the empirical training data distribution $\hat{p}_{\text{data}}(G)$ and $\hat{p}_{\text{data}}(O|G)$. By  defining $\mathbf{X} \eqdef (\mathbf{A}, \mathbf{W}, \mathbf{F}, \ell)$, our model can be written as
\begin{equation}
p_\theta(G,O,\mathbf{Z},\mathbf{X}) = \hat{p}_{data}(G) \; \hat{p}_{data}(O|G) \; p_{\theta_\text{FM}}(\mathbf{Z}|G, O) \; p_{\theta_\mathcal{D}}(\mathbf{X}|\mathbf{Z},G,O),
\end{equation}
where $\theta = (\theta_\text{FM}, \theta_\mathcal{D})$ are the parameters of the flow matching and decoder neural networks. Similar to ADiT, we use a Diffusion Transformer (DiT) \cite{dit} as our denoiser network $\mathcal{F}$, where conditional information is incorporated through adaptive layer norm. Moreover, we use self-conditioning \cite{self_cond}, in which the denoiser's prediction from the previous timestep is concatenated with the current input, with a 50\% dropout probability during training. We condition on the space group via classifier-free guidance (CFG) \cite{cfg} to incorporate the conditioning on space group into the latent generative process (see Appendix \ref{app:space_group_cond} for an ablation study on this conditioning).\footnote{This is also implemented in ADiT, although it is not explicitly described in the paper.} At each denoising step, the model produces two predictions: one conditioned on the space group and one without. These two predictions are combined using guidance scale $\gamma$ as a weighted combination, controlling the influence of the space group conditioning. Ablation studies on self-conditioning and CFG scale are provided in Appendix \ref{app:ablation_studies}. Consistent with prior work \cite{wyckoff_diff, sgequi_diff}, we jointly sample Wyckoff positions and atomic species, ensuring that Wyckoff positions with zero degrees of freedom are not sampled more than once for different orbits, as this would result in overlapping atomic positions.  Pseudocode for generation is presented in Algorithm \ref{alg:generation}.

\begin{figure}[t]
\begin{center}
\begin{minipage}{0.7\linewidth}
\begin{algorithm}[H]
\footnotesize
\caption{Pseudocode for DiT sampling}
\label{alg:generation}
\begin{algorithmic}[1]

\Statex \hspace*{-\algorithmicindent}\textbf{Input:} $G$, integration steps $T$, cfg. scale $\gamma$
\Statex \hspace*{-\algorithmicindent}\textbf{Output:} Generated sample $(\mathbf{A}, \mathbf{W}, \mathbf{F}, \ell)$

\Statex \textcolor{orange}{\# Sample priors and initial noisy latents $Z^{(0)}$ at $t=0$}
\State Sample $G \sim \hat{p}_{\text{data}}(G)$
\State Sample $O \sim \hat{p}_{\text{data}}(O \mid G)$ \quad \textcolor{orange}{\# orbit (index set)}
\State $\mathbf{Z}_O^{(0)} = \{z_i^{(0)} \sim \mathcal{N}(0,1)^d : i \in O\}$
\State $\Delta t = 1 / T$  

\Statex \textcolor{orange}{\# Denoising loop for orbit $O$}
\State $\text{for } t \in \mathrm{linspace}(0.0, 1.0, T):$
\State \quad $\mathbf{Z}'_{O,\mathrm{cond}} = \mathcal{F}(\mathbf{Z}_O^{(t)}, t, G)$ 
\hspace{2em} \textcolor{orange}{\textit{\# Conditional prediction}}

\State \quad $\mathbf{Z}'_{O,\mathrm{uncond}} = \mathcal{F}(\mathbf{Z}_O^{(t)}, t, \phi)$ 
\hspace{2em} \textcolor{orange}{\textit{\# Unconditional prediction}}

\State \quad $\mathbf{Z}'_O = (1 - \gamma)\, \mathbf{Z}'_{O,\mathrm{uncond}} + \gamma\, \mathbf{Z}'_{O,\mathrm{cond}}$ \hspace{2em} \textcolor{orange}{\textit{\# Classifier-free guidance}}

\State \quad $\mathbf{Z}_O^{(t+\Delta t)} = \mathbf{Z}_O^{(t)} + \Delta t \cdot \frac{\mathbf{Z}'_O - \mathbf{Z}_O^{(t)}}{1 - t}$ \hspace{2em} \textcolor{orange}{\textit{\# Euler step}}

\State $(\mathbf{A}, \mathbf{W}, \mathbf{F}, \ell) = \mathcal{D}(\mathbf{Z}^{(1)}, G)$ \hspace{2em}\textcolor{orange}{\# Decode latents to ASU (Algorithm 2)}

\State $C = \text{PyXtal}(G, \mathbf{W}, \mathbf{A}, \mathbf{F}, \ell)$  \hspace{2em}\textcolor{orange}{\# Construct crystal}

\end{algorithmic}
\end{algorithm}
\end{minipage}
\end{center}
\end{figure}

\section{Related Work}
\label{sec:related_works}
\subsection{Symmetry-agnostic approaches}
Symmetry-agnostic methods do not explicitly enforce crystallographic symmetries, but instead rely on learning them from data. CDVAE \cite{cdvae} is an early example that applies diffusion to generate atom coordinates, while a variational autoencoder models the unit cell shape. DiffCSP \cite{diffcsp} extends on CDVAE, defining a joint diffusion process over atoms and unit cell. MatterGen \cite{MatterGen2025} is similar to DiffCSP, but also allows conditioning on properties. The authors investigated conditioning on space groups, but still, in their experiments only 58 \% of the generated structures are classified as the space group they were conditioned on. FlowMM \cite{flowmm} is a model based on flow matching that extends flow matching to respect symmetries such as translation, rotation, permutation, and periodic boundaries, but does not consider the types of symmetry we do in our work. The model we build on, ADiT \cite{adit}, forms a unified generative framework that generates both periodic crystals and non-periodic molecular systems. We focus exclusively on crystal generation, as the crystallographic symmetries considered in this work are specific to periodic structures. CrystalDiT \cite{crystal_dit}, similarly to ADiT, uses a Diffusion Transformer \cite{dit} approach, but performs its diffusion in data space. OMatG \cite{omatg} unifies diffusion and flow matching through stochastic interpolants, jointly generating atomic positions and lattice vectors alongside discrete flow matching for atomic species.
\subsection{Symmetry-aware approaches}
DiffCSP++ \cite{diffcsp++} is an early symmetry-aware method: an extension of DiffCSP where generation starts with choosing a \emph{template}, i.e., space group and occupied Wyckoff positions, from the training set. Generation then proceeds by generating elements, atomic positions, and unit cell similar to DiffCSP, while respecting the symmetries determined by the template. SymmCD \cite{symmcd} is a method that, similar to SymADiT, focuses on modelling of the ASU. However, its generative procedure still operates on and processes the full expanded material. WyCryst \cite{wycryst} and WyckoffDiff \cite{wyckoff_diff} take a similar approach in that they generate the symmetry properties of crystals (space group, Wyckoff positions, elements), but they do not model the remaining degrees of freedom, while we generate the complete geometry. SGEquiDiff \cite{sgequi_diff} uses a multi-stage approach, where the space group and unit cell are first generated, and an autoregressive model then generates the occupied Wyckoff positions and elements. Conditioned on this information, a space group-equivariant diffusion process generates the atomic positions, respecting the geometric constraints enforced by the Wyckoff positions. 

\section{Experiments and Results}
 In this section, we present SymADiT model details, datasets, evaluation pipeline, baselines used for comparison, and experimental results.
\label{sec:experiments_and_results}
\paragraph{Training and Hyperparameters}
Similar to ADiT, the encoder $\mathcal{E}$ and decoder $\mathcal{D}$ use a standard Transformer architecture. Unlike ADiT, however, we do not use positional embeddings. We use an embedding dimension of $d_m = 512$, a latent space dimension of $d = 32$, 8 attention heads, and 8 layers, for a total of 52M parameters. The second-stage DiT denoiser, with the same latent dimension and 12 heads and layers, contains 130M parameters. Due to the nature of the Transformer being easily scalable, we also explore increasing the DiT to 450M parameters ($\text{DiT}_L$), similar to ADiT. For the generation, we use cfg scale $\gamma=2$ (focusing more on the conditional path), with the number of time steps $T = 500$. Models are trained for a maximum of \thsnd{7} epochs, with batch size 512, on either A100 (80GB) or H200 (141GB) GPUs. 
\paragraph{Datasets}
We evaluate our model on three datasets. The first is the commonly used MP20 dataset \cite{cdvae}, which contains $45{,}229$ crystals with a maximum of 20 atoms in the primitive unit cell. For a fair comparison, we use the same split used in the original paper, meaning $27{,}138$ crystals are used for training, and $9{,}046$ for validation, with the remaining reserved for testing (which are not used). In addition, we evaluate our own models on the MPTS52 \cite{mpts52} dataset, containing $40{,}476$ experimentally verified crystals with a maximum of 52 atoms in the primitive unit cell. The training split contains $27{,}380$ crystals, with approximately $32\%$ overlap with the MP20 training set. Furthermore, we explore merging the two datasets (MP20+MPTS52), creating a dataset with $65{,}886$ crystals. We divide the data into training and validation sets using an $80/20$ split, ensuring similar space group distributions across both sets.
\paragraph{Evaluation} Following the literature, we evaluate all methods on \thsnd{10} generated crystals and report structural and compositional validity. To provide a more comprehensive assessment of the generated crystals, we additionally use the following evaluation pipeline: Of the \thsnd{10} sampled crystals, we filter out crystals that do not pass the \emph{structural validity} check.\footnote{As others have noted \cite{sgequi_diff, wyckoff_transformer}, only $\sim 90\%$ of the MP20 training set passes the compositional validity check based on SMACT \cite{smact}, and we do therefore not filter based on this criterion.} We then compute \emph{uniqueness} (U) on those crystals, which compares the generated crystals to each other to check if they are duplicates. Following that, we sample \thsnd{1} crystals from those unique and structurally valid crystals, to calculate \emph{novelty} (N), which measures whether the unique crystals are present in the training data. Both uniqueness and novelty are checked through \texttt{StructureMatcher} \cite{pymatgen} with \texttt{stol}=$0.3$, \texttt{angle\_tol}=$10.0$ and \texttt{ltol}=$0.2$, following ADiT \cite{adit}. These are reported as conditional probabilities.\footnote{E.g., structural validity = $9500/10000$, uniqueness = $8500/9500$, novelty = $500/1000$.} 
Finally, we randomly select $500$ structurally valid, unique, and novel ($\text{U} \mid \text{N}$) structures and perform expensive Density Functional Theory (DFT) \cite{hohenberg1964inhomogeneous, kohn1965self} calculations to assess thermodynamic stability (S) and metastability (M.S). This forms the S.U.N. and M.S.U.N. metrics, for which we report how many of the 500 structures satisfy the corresponding DFT-based stability criteria. We follow a similar DFT evaluation as others \cite{adit, sgequi_diff, flowmm}, with details presented in Appendix \ref{app:dft_details} for reproducibility. 

To further assess how well the methods capture the underlying symmetry in the data, we investigate the fraction of generated crystals belonging to space group P1. These structures exhibit only translational symmetry. Moreover, we evaluate distributional properties by reporting Jensen–Shannon divergences (JSD) for both the space group ($\text{JSD}_{\text{G}}$) and Wyckoff positions ($\text{JSD}_{\text{Wy}}$) distributions. For the Wyckoff analysis, divergences are averaged across space groups, weighted by the number of generated samples in each group. Moreover, we report the Wasserstein  distance for the distribution of the number of atoms in the unit cell $W_{\text{A}}$. 
\paragraph{Metric applicability.}
\label{par:metric_applicability}
Some reported metrics are not applicable to all methods. In particular, P1 and $\mathrm{JSD}_{\text{G}}$ are not applicable to symmetry-aware methods, since these methods all begin their generation pipeline by sampling the space group directly from the empirical training distribution.\footnote{MP20 contains 2.23\% P1 structures.} Similarly, $W_A$ is not applicable to the symmetry-agnostic methods or to DiffCSP++, since these methods sample the number of atoms from the empirical training distribution. Appendix~\ref{app:eval_metrics} contains all details on the evaluation metrics.
\paragraph{Baselines} On MP20, we compare our model to both symmetry-aware and symmetry-agnostic approaches. The symmetry-aware methods include SGEquiDiff \cite{sgequi_diff}, SymmCD \cite{symmcd}, and DiffCSP++ \cite{diffcsp++}. For symmetry-agnostic methods, we include ADiT \cite{adit}, CrystalDiT \cite{crystal_dit}, MatterGen \cite{MatterGen2025}, FlowMM \cite{flowmm}, DiffCSP \cite{diffcsp}, and OMatG \cite{omatg}. For ADiT, SymmCD, SGEquiDiff, and OMatG, we trained the models from scratch and ran their generation pipelines, with no modifications made to any of the codebases. The remaining crystals were obtained through CrystalDiT \cite{crystal_dit}.\footnote{CIF files can be found on HuggingFace: \url{https://huggingface.co/xiaohan-yi/CrystalDiT}.} See Appendix \ref{app:licenses} for Licenses. 
\begin{table}[t]
\centering
\footnotesize 	
\caption{\textbf{MP20 dataset.} Metrics are reported over \thsnd{10} generated crystals. Validity is evaluated on all samples. Uniqueness is measured as the fraction of structurally valid samples that are distinct from each other. Novelty is computed over \thsnd{1} unique structures to assess duplication with the training data. (Meta)stability is evaluated by DFT on 500 samples that are structurally valid, unique, and novel. The M/S.U.N columns report the number of (meta)stable structures among the 500 samples evaluated with DFT. Dash indicates that the method uses the empirical training distribution and the metric is therefore not meaningful.} 
\label{tab:mp20_results}

\begin{adjustbox}{max width=\linewidth}
\begin{tabular}{c c c cc ccc c c c}
\toprule
Method & Sym.
& \multicolumn{2}{c}{Validity (\%) $\uparrow$}
& \multicolumn{3}{c}{Distr. distance $\downarrow$}
& \% P1
& \% $\text{U} \mid \text{N}$ $\uparrow$
& \# \text{S.U.N.} $\uparrow$
& \# \text{M.S.U.N.} $\uparrow$ \\
\cmidrule(lr){3-4}
\cmidrule(lr){5-7}
& & Struct. & Comp.
& $\mathrm{JSD}_G$
& $\mathrm{JSD}_{\text{Wy}}$
& $W_{\text{A}}$
& & & & \\
\midrule

SGEquiDiff \cite{sgequi_diff}
& Yes
& 99.65
& 84.56
& -- %
& 0.05
& 0.48
& -- %
& $98.35 \mid 89.50$
& 51 
& 89 \\ 

SymmCD \cite{symmcd}
& Yes
& 92.74
& 81.53
& -- %
& 0.13
& 1.17
& -- %
& $98.63 \mid 92.70$
& 49
& 78 \\ 

DiffCSP++ \cite{diffcsp++}
& Yes
& 99.96
& 84.48
& -- %
& 0.04
& --
& -- %
& $98.27 \mid 89.80$
& 64 
& 105\\ 

\midrule

ADiT \cite{adit}
& No
& 99.56 
& 90.44
& 0.95
& 0.70
& -- %
& 99.01
& $88.23 \mid 48.60$
& 103
& 170 \\

CrystalDiT \cite{crystal_dit}
& No
& 97.82
& 84.34
& 0.94
& 0.67
& -- %
& 98.25
& $92.10 \mid 65.10$
& 47
& 152 \\

MatterGen \cite{MatterGen2025}
& No
& 100.00
& 83.58
& 0.81
& 0.32
& -- %
& 67.12
& $98.37 \mid 91.80$
& 41 
& 126 \\

FlowMM \cite{flowmm}
& No
& 99.23
& 82.19
& 0.91
& 0.35
& -- %
& 90.93
& $98.52 \mid 90.90$
& 36
& 106 \\

DiffCSP \cite{diffcsp}
& No
& 99.93
& 81.94
& 0.85
& 0.40
& -- %
& 70.60
& $97.35 \mid 91.10$
& 53 
& 107 \\

OMatG \cite{omatg}
& No
& 99.13
& 94.52
& 0.95
& 0.55
& -- %
& 99.20
& $97.62 \mid 80.10$
& 87 
& 134 \\

\midrule

SymADiT
& Yes
& 96.50
& 88.23
& -- %
& 0.04
& 0.22
& -- %
& $87.42 \mid 54.60$ %
& 85
& 136 \\

\bottomrule
\end{tabular}
\end{adjustbox}
\end{table}

\paragraph{MP20 Results}
Table \ref{tab:mp20_results} compares SymADiT to both symmetry-aware and symmetry-agnostic approaches, showing competitive performance. First off, it is evident that the symmetry-agnostic methods struggle with capturing the symmetry in the data: The fraction of P1 crystals (P1 column) in MP20 is $2.23 \%$, but the symmetry-agnostic model closest to this is MatterGen with $67 \%$, and for ADiT\footnote{While ADiT is listed as symmetry-agnostic, it tries to incorporate symmetry via conditioning on space group during generation, but this does not seem to be effective.} and CrystalDiT, which are the symmetry-agnostic methods most related to SymADiT, this number is $>98 \%$. Appendix \ref{app:gen_sg_dist} showcases further the proportion of space group symmetries for each method and highlights more clearly how symmetry-agnostic methods in general fail to capture the space group distribution in relation to MP20 training data. Furthermore, SymADiT most closely follows the training distribution among the symmetry-aware methods. This suggests that, for some competing symmetry-aware methods, the final generated structures may not always preserve the space group initially sampled during generation, thereby altering the resulting space group distribution.

The distribution metrics in Table~\ref{tab:mp20_results} further highlight that symmetry-agnostic methods are considerably less effective at capturing the symmetry characteristics of the data in terms of Wyckoff positions ($\text{JSD}_\text{Wy}$). An interesting observation is that symmetry-aware methods only sample implicitly from the atom distribution ($\text{W}_{A}$) when sampling the occupied Wyckoff positions, but still manage to capture this distribution well. In Appendix \ref{app:additional_results}, we show additional statistical properties of novel crystals generated by each method.

\emph{Taken together, these results highlight the need for models like SymADiT that incorporate symmetry as an inductive bias, rather than learning it purely from data}. Furthermore, the results suggest that the inductive bias incorporated into the representation is sufficient to generate symmetric crystals, without requiring additional equivariance constraints in the generative model: \emph{a simple Transformer combined with latent flow matching is sufficient}.

Investigating the \morsun columns in Table \ref{tab:mp20_results}, we note that \emph{SymADiT is able to achieve the highest number of \morsun crystals among all symmetry-aware methods}, showcasing its strong performance. SymADiT does not, by comparison, achieve higher \morsun numbers than ADiT. However, as stated and shown above, symmetry-agnostic methods struggle to capture the underlying symmetry in the data, neglecting a key characteristic of realistic and synthesizable crystals: \emph{most experimentally observed crystals exhibit substantial symmetry and do not belong to the P1 space group.}

Thus, while the \morsun metric is an important measure of performance at this stage of the crystal generation pipeline, it is not the ultimate measure of success. The final goal is the experimental synthesis of the proposed crystals, where SymADiT (and other symmetry-aware methods) may provide a more suitable foundation by explicitly capturing the symmetry characteristics commonly observed in experimentally realized crystals. A statistical comparison of SymADiT's \morsun performance against other methods is presented in Appendix~\ref{app:statistical_analysis}.

The uniqueness and novelty fractions of crystals generated by SymADiT and ADiT are both comparably low. This suggests that the latent-space design could be the contributing factor. One possible interpretation is that the lower-dimensional latent space used by SymADiT and ADiT makes memorization or concentration around the training manifold more likely. Such a bias toward the training distribution may also help explain the strong \morsun performance of ADiT and SymADiT. Samples generated closer to the training manifold may be more likely to remain physically realistic and stable, suggesting that the latent-space bottleneck may trade some novelty for improved structural realism. 

From a practical perspective, a lower uniqueness and novelty simply mean that one would have to generate more crystals to reach a desired number of crystals to be evaluated. Since generative models are computationally cheap compared to DFT, this extra sampling cost is not a major concern in practice, and it follows that what matters most is the quality of the crystals that make it to the final evaluation stage (DFT), with the previous steps, like novelty, uniqueness, and structural validity, acting as filters. In this regard, SymADiT is the best among the symmetry-aware methods. This does not, of course, mean that these metrics are unnecessary, but rather that they require a more nuanced interpretation. We further discuss the novelty metric in more detail in Appendix \ref{app:novelty_discussion}.

\begin{table*}[t]
\centering
\footnotesize 	
\caption{\textbf{MPTS52 and MP20+MPTS52.} Metrics are reported over \thsnd{10} generated crystals, with \morsun computed using 100 DFT calculations. $\mathrm{SymADiT}_{L}$ refers to the larger 450M DiT model. See caption of Table~\ref{tab:mp20_results} for additional details.}
\label{tab:mp20_and_mpts52_results}

\begin{adjustbox}{max width=\linewidth}
\begin{tabular}{l c cc ccc c c c c}
\toprule
Method
& Data
& \multicolumn{2}{c}{Validity (\%) $\uparrow$}
& \multicolumn{3}{c}{Distr. distance $\downarrow$}
& $\%$ P1
& \% $\text{U} \mid \text{N}$ $\uparrow$
& \# \text{S.U.N.} $\uparrow$
& \# \text{M.S.U.N.} $\uparrow$ \\
\cmidrule(lr){3-4}
\cmidrule(lr){5-7}

& & Struct.
& Comp.
& $\mathrm{JSD}_G$
& $\mathrm{JSD}_{\text{Wy}}$
& $W_{\text{A}}$
& & & & \\
\midrule
SymADiT
& MPTS52
& 94.34
& 84.23
& -- %
& 0.04
& 0.40
& -- %
& $85.04 \mid 48.50$
& 8 %
& 20 \\ %

SymADiT
& MP20 + MPTS52
& 95.31
& 86.60
& -- %
& 0.04
& 0.32
& -- %
& $93.94 \mid 54.30$
& 9 %
& 30 \\ %

$\mathrm{SymADiT}_{L}$
& MP20 + MPTS52
& 94.96
& 87.27
& -- %
& 0.05
& 0.49
& -- %
& $92.57 \mid 49.50$
& 7 %
& 29 \\ %
\bottomrule
\end{tabular}
\end{adjustbox}
\end{table*}

\paragraph{Comparison of Inference Speed between SymADiT and ADiT}
By operating on symmetry orbits rather than individual atoms, SymADiT substantially reduces the number of tokens processed by the Transformer compared with ADiT, resulting in a more compact representation and faster generation. Specifically, on the MP20 dataset, SymADiT processes an average of $4.74$ tokens per sample, compared with $10.34$ for ADiT, resulting in a $2.40\times$ inference speed-up over ADiT. Further details about how the experiment was conducted are provided in Appendix~\ref{app:inference_speed}.

\paragraph{Additional Results}
Table \ref{tab:mp20_and_mpts52_results} shows additional results for SymADiT on larger datasets. On both datasets, due to computational constraints, we ran only 100 DFT calculations on these models. SymADiT still performs reasonably in the distribution metrics, showing that the model can robustly scale up to larger structures. We found that the larger DiT did not improve the generation of novel structures, indicating that the 130M model is sufficient. %
Visual results of \morsun structures generated by SymADiT are presented in Appendix \ref{app:visual_results}, capturing a wide range of space groups.

\paragraph{Ablation Studies}
Ablation studies on AE latent regularisation as well as CFG and self-conditioning for the DiT are provided in Appendix \ref{app:ablation_studies}.

\section{Discussion and Conclusion}
\label{sec:conclusion}
In this paper, we propose SymADiT, a symmetry-aware extension of ADiT, a generative framework for crystals based on latent flow matching. We construct SymADiT using a Wyckoff representation that explicitly encodes symmetry. Our results demonstrate the capability of SymADiT to generate crystals with realistic symmetry properties, which ADiT and other symmetry-agnostic methods struggle to capture, while achieving the highest number of thermodynamically stable crystals among the symmetry-aware methods evaluated. Notably, these improvements are achieved using a simple off-the-shelf Transformer architecture, without requiring additional architectural modifications, while also reducing the number of tokens relative to ADiT.

Due to computational constraints, we conducted only 100 DFT evaluations for Table \ref{tab:mp20_and_mpts52_results}, adding unknown variance to those reported \morsun rates. 
A limitation of the generative model developed here and in previous work is that, while proposing new crystals, it provides no guidance on how they could be synthesized, which leaves an open question for future work. Additionally, while ADiT was originally developed for enabling generation of both crystals and molecules, we focused only on the crystal generation aspect, and future work could include molecule generation capability together with our symmetry-aware approach to further study transfer learning between the two atomic systems.

\section*{Acknowledgements}%
This work was supported by the Knut and Alice Wallenberg Foundation (KAW) via the Wallenberg AI, Autonomous Systems and Software Program (WASP) and the Wallenberg Initiative Material Science for Sustainability (WISE) through the joint WASP-WISE project \textit{Accelerated design of functional 2D materials by leveraging foundation models in a closed-loop approach}, and by Excellence Center at Linköping--Lund in Information Technology (ELLIIT). Computations were enabled by the Berzelius resource provided by the Knut and Alice Wallenberg Foundation at the National Supercomputer Centre (NSC) and by resources provided by the National Academic Infrastructure for Supercomputing in Sweden (NAISS) at the National Supercomputer Centre (NSC), partially funded by the Swedish Research Council through grant agreement no. 2022-06725.

\bibliographystyle{unsrtnat}
\bibliography{references}

\newpage
\appendix
\section{Example of the Representation}
\label{app:example_of_rep}
Here we showcase an example of how our representation can be used to create a crystal $\mathbf{C} = (G, \mathbf{W}, \mathbf{A}, \mathbf{F}, \ell) $ using the crystal library \texttt{PyXtal} \cite{pyxtal}.  We take the classic example of NaCl, which belongs to space group $225$. Na and Cl occupy two highly symmetric orbits in space group $225$, and have Wyckoff positions "4a" and "4b", respectively, which results in 8 atoms once expanded to the full conventional unit cell. The representatives of these orbits are completely fixed by symmetry: $(0,0,0)$ for Na at "4a", and $(0.5, 0.5, 0.5)$ for Cl at "4b" (i.e., both representatives have 0 DOF), meaning that no prediction of  $\mathbf{F}$ is required. Furthermore, the unit cell in space group $225$ is a cube, meaning that all sides have the same length (in the case of NaCl, $5.65\ \text{Å}$), with angles all being fixed at $90 \degree$. Again, this implies that the autoencoder only gets supervised with one number for the unit cell, namely the length of the cube ($a$ in $\ell$). \texttt{PyXtal} combines the free fractional and free unit cell parameters into a tuple $\mathbf{X}$, which in this case would be $\mathbf{X} = (5.65)$. Given all of this, the tuple needed to build the crystal via \texttt{PyXtal} would look like      

\begin{equation*}
\setlength{\abovedisplayskip}{4pt}
\setlength{\belowdisplayskip}{4pt}
\mathbf{C} = (G, \mathbf{W}, \mathbf{A}, \mathbf{F}, \ell)
= (G, \mathbf{W}, \mathbf{X}, \mathbf{A})
= [225,\; (4a, 4b),\; (5.65),\; (\text{Na}, \text{Cl})].
\end{equation*}
This can later be expanded to the full crystal using pymatgen \cite{pymatgen}.

\paragraph{Generation}
From a generative standpoint, the process of generating NaCl would begin by sampling space group 225 from the empirical training distribution. Next, Wyckoff positions and atomic species are sampled, with the constraint that zero-degree-of-freedom (0 DOF) Wyckoff positions cannot be selected more than once (e.g., assigning “4a” to multiple orbits is disallowed). Since both “4a” and “4b” are selected, positional predictions are not utilized (no free parameters). Similarly, for lattice parameter prediction, only one degree of freedom is available (length of cube); therefore, only the first prediction is retained, and all subsequent predictions of $\ell$ are discarded.

\section{Ablation Studies}
\label{app:ablation_studies}
We conduct extensive ablation studies to justify our choice of training settings.

To assess latent-space regularisation, we compare saturation-based regularisation with KL regularisation at different scales, as well as using no latent space regularisation at all. Reconstruction performance for stage 1 is evaluated using the match rate (MR) and root mean square distance (RMSD) on the MP20 test set. Overall, we found that using a downweighted KL divergence penalty did not have any significant effect on performance, which is why we decided to follow \cite{lost_in_latent_space} and use a saturation function instead. Moreover, we note that we achieve an approximately 7\% increase in match rate compared to ADiT's reconstruction performance, as well as a lower RMSD \cite{adit}. Results are presented in Table \ref{tab:regularisation_ablation}.

\begin{table}[h]
\centering
\footnotesize
\caption{\textbf{Regularisation ablation.} Effect of different regularisation methods on match rate (MR) and RMSD.}
\label{tab:regularisation_ablation}
\setlength{\tabcolsep}{6.0pt}
\renewcommand{\arraystretch}{1.30}

\begin{tabular}{l cc}
\toprule
Reg. method & MR $\uparrow$ & RMSD $\downarrow$ \\
\midrule
No regularisation
& 97.92
& 0.029 \\

Saturation
& 97.90
& \textbf{0.019} \\

KL scale = 0.1
& 79.98
& 0.104 \\

KL scale = 0.01
& 95.35
& 0.040 \\

KL scale = 0.001
& 97.79
& 0.022 \\

KL scale = 0.0001
& \textbf{98.93}
& \textbf{0.019} \\
\bottomrule
\end{tabular}
\end{table}

    To assess the effect of classifier-free guidance (CFG) during generation, we vary the CFG scale and evaluate generation performance using uniqueness and novelty. Higher values indicate better performance. Lower CFG scales (focusing less on the space group conditioning) achieve higher novelty at the cost of validity metrics. Overall, we found that CFG scales of 1.5 and 2.0 work well in our experiments, placing greater emphasis on conditioning. Results are presented in Table \ref{tab:cfg_ablation}
\begin{table}[h]
\centering
\footnotesize
\caption{\textbf{Classifier-free guidance ablation.} Effect of the CFG scale on structural validity, compositional validity, uniqueness, and novelty.}
\label{tab:cfg_ablation}
\setlength{\tabcolsep}{5.0pt}
\renewcommand{\arraystretch}{1.30}

\begin{tabular}{c cccc}
\toprule
CFG
& Struct. (\%) $\uparrow$
& Comp. (\%) $\uparrow$
& U (\%) $\uparrow$
& N (\%) $\uparrow$ \\
\midrule
0.0
& 70.03
& 65.83
& \textbf{99.73}
& \textbf{96.50} \\

0.5
& 92.48
& 82.62
& 97.12
& 74.90 \\

1.0
& 93.20
& 83.25
& 82.40
& 48.20 \\

1.5
& 96.86
& 88.18
& 90.72
& 54.40 \\

2.0
& 96.50
& 88.23
& 87.42
& 54.60 \\

2.5
& \textbf{97.14}
& 88.27
& 87.69
& 52.40 \\

3.0
& 97.05
& \textbf{88.41}
& 86.34
& 54.00 \\
\bottomrule
\end{tabular}
\end{table}
.

To assess the effect of self-conditioning during training, we conduct an ablation in which the self-conditioning probability is set to zero. We train an otherwise identical model using the same experimental setup described in Table 1 in the paper. Without self-conditioning, higher uniqueness is achieved, although with lower novelty. Results are presented in Table \ref{tab:self_conditioning_ablation}
\begin{table}[h]
\centering
\footnotesize
\caption{\textbf{Self-conditioning ablation.} Effect of self-conditioning on structural validity, compositional validity, uniqueness, and novelty.}
\label{tab:self_conditioning_ablation}
\setlength{\tabcolsep}{5.0pt}
\renewcommand{\arraystretch}{1.30}

\begin{tabular}{l cccc}
\toprule
Self-cond.
& Struct. (\%) $\uparrow$
& Comp. (\%) $\uparrow$
& U (\%) $\uparrow$
& N (\%) $\uparrow$ \\
\midrule
Yes
& 96.50
& 88.23
& 87.42
& \textbf{54.60} \\

No
& \textbf{97.85}
& \textbf{88.27}
& \textbf{89.20}
& 50.40 \\
\bottomrule
\end{tabular}
\end{table}
.

\section{Network Details}
\label{app:network_and_training}
\paragraph{Autoencoder}
The MLPs in the encoder (Algorithm \ref{alg:ae_encoder}) for the fractional coordinates $\mathbf{F}$ and the unit cell $\ell$ are two-layer feedforward networks, using a SiLU \cite{silu} nonlinearity between layers. We concatenated $\mathbf{F}$ and $\ell$ with their corresponding mask $M_\mathbf{F}$ and $M_\ell$ before passing to the MLP. For the atom types $\mathbf{A}$ and space group $G$, we use a typical \texttt{nn.Embedding} from PyTorch \cite{pytorch}. For the Wyckoff positions $\mathbf{W}$, since Wyckoff letters and multiplicities (e.g., \texttt{4a}, \texttt{4b}) are not unique across different space groups, we encode each position by concatenating its space group identifier with the Wyckoff label (e.g., \texttt{225\_4a}). Again, we use an \texttt{nn.Embedding} layer to encode this. The Linear layers in the decoder (Algorithm \ref{alg:ae_decoder}) are single-layer neural networks for the discrete predictions ($\mathbf{W}$ and $\mathbf{A}$), and similar two-layer networks as the encoder for the continuous predictions ($\mathbf{F}$ and $\mathbf{\ell}$).  

During training, the free parameters of $\mathbf{F}$ (constrained by $G$ and $\mathbf{W}$, see \nameref{par:wyckoff}) and $\ell$ (constrained by $G$, see \nameref{par:unit_cell}) are indicated by binary masks $M_{\mathbf{F}} \in \{0,1\}^3$ and $M_{\ell} \in \{0,1\}^6$. For Wyckoff positions, invalid entries are masked by assigning their logits to $-\infty$ prior to the softmax operation.

\section{Latent Space Visualisation of Autoencoder}
\label{app:latent_viz}
\begin{figure}[H]
    \centering
    \includegraphics[width=0.5\textwidth]{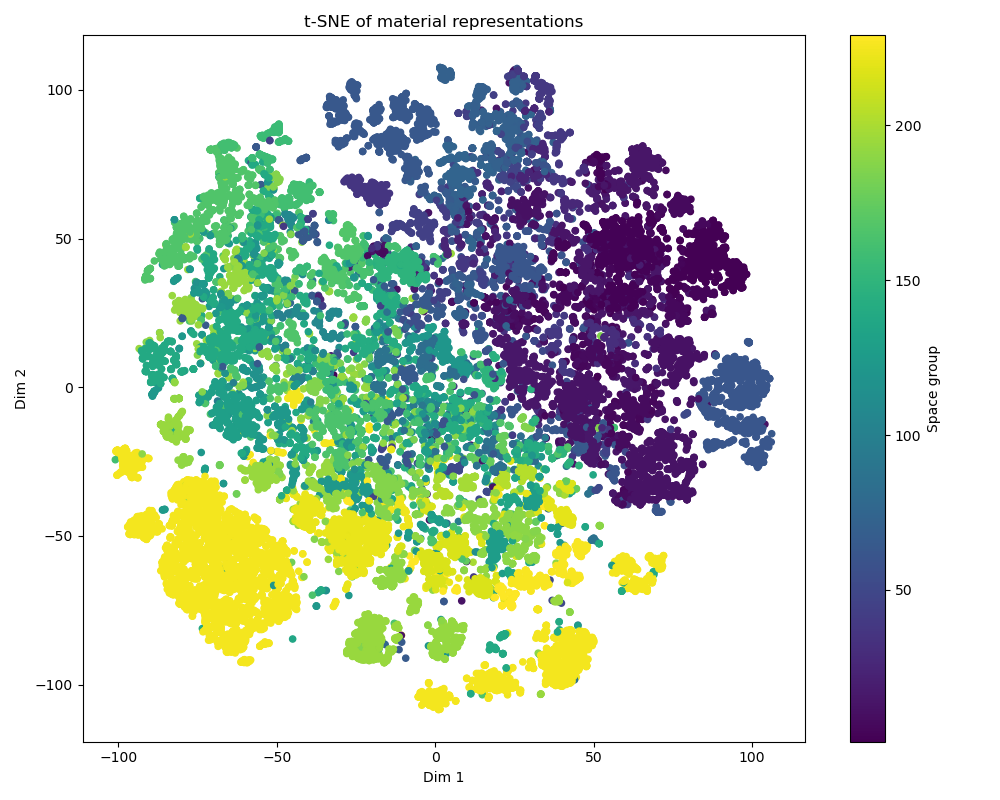}
    \caption{t-SNE and PCA plot of latent embeddings from a trained autoencoder on the MP20 dataset.}
    \label{fig:latent_viz}
\end{figure}

We project the learned crystal representations to two dimensions using t-SNE, preceded by PCA for dimensionality reduction and denoising. Figure \ref{fig:latent_viz} shows this for \thsnd{30} samples for the MP20 dataset, with the colours indicating which space group each crystal belongs to. We observe that the latent representations form distinct clusters according to the crystals’ space groups, with similar space groups located closer to one another. Still, we observe some mixing, which could be due to other information that the crystals share (e.g. similar Wyckoff or unit cell information). We observe that removing the space group conditioning during the latent flow matching part results in fewer valid crystals (see Appendix \ref{app:space_group_cond}) and hypothesize that this additional information helps guide the noisy latents toward the correct cluster.

We evaluated several strategies for encoding local information in Algorithm \ref{alg:ae_encoder}. In particular, we experimented with concatenating and then down‑projecting local, global, and combined representations. Ultimately, we found that a simple additive combination yielded the best performance in terms of generating novel and stable crystals.

\section{Space Group Conditioning}
\label{app:space_group_cond}
\begin{figure}[H]

    \centering
    \begin{subfigure}{0.48\textwidth}
        \centering
        \includegraphics[width=\linewidth]{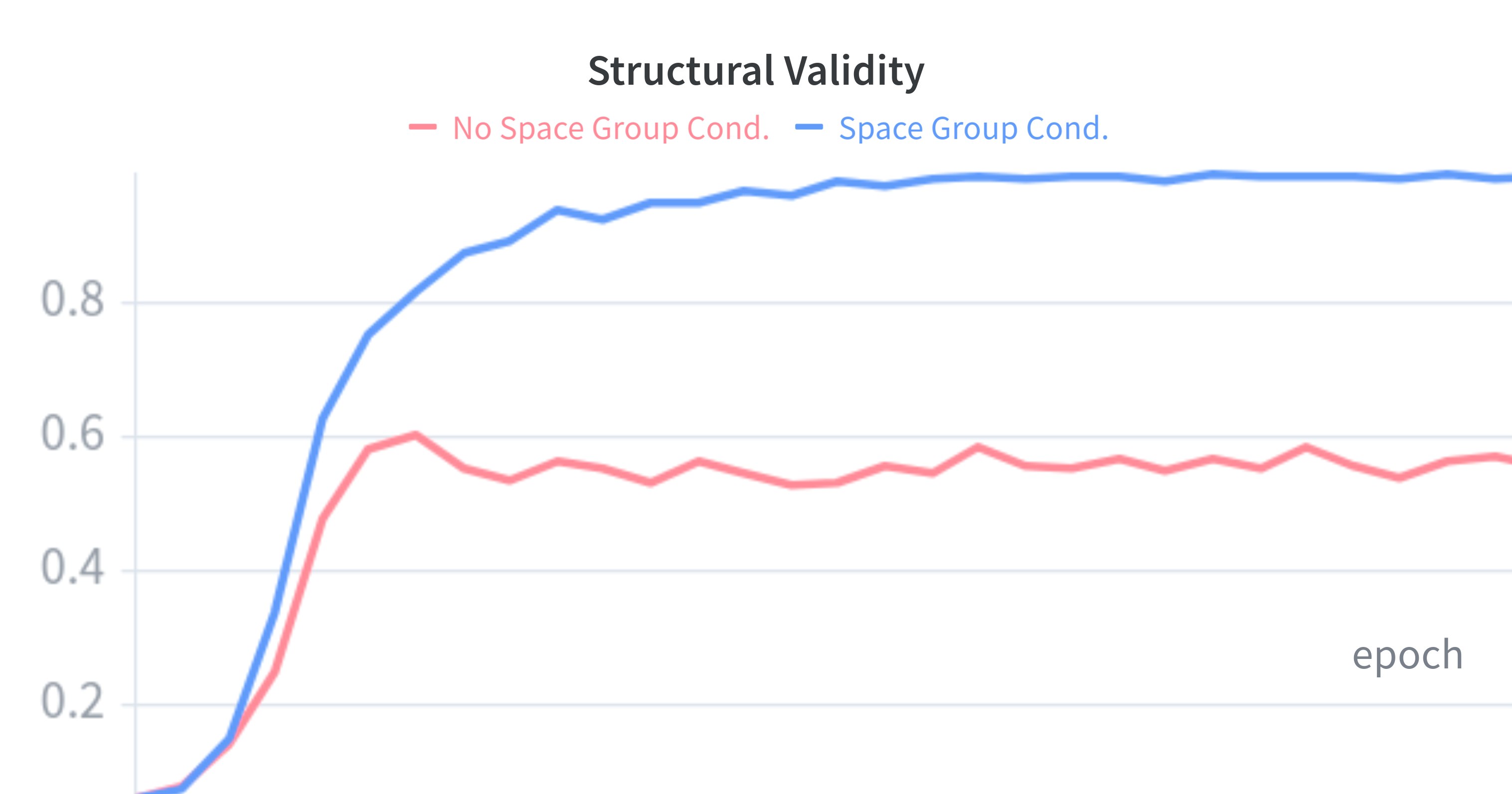}
        \caption{Structural Validity}
        \label{fig:no_sg}
    \end{subfigure}
    \hfill
    \begin{subfigure}{0.48\textwidth}
        \centering
        \includegraphics[width=\linewidth]{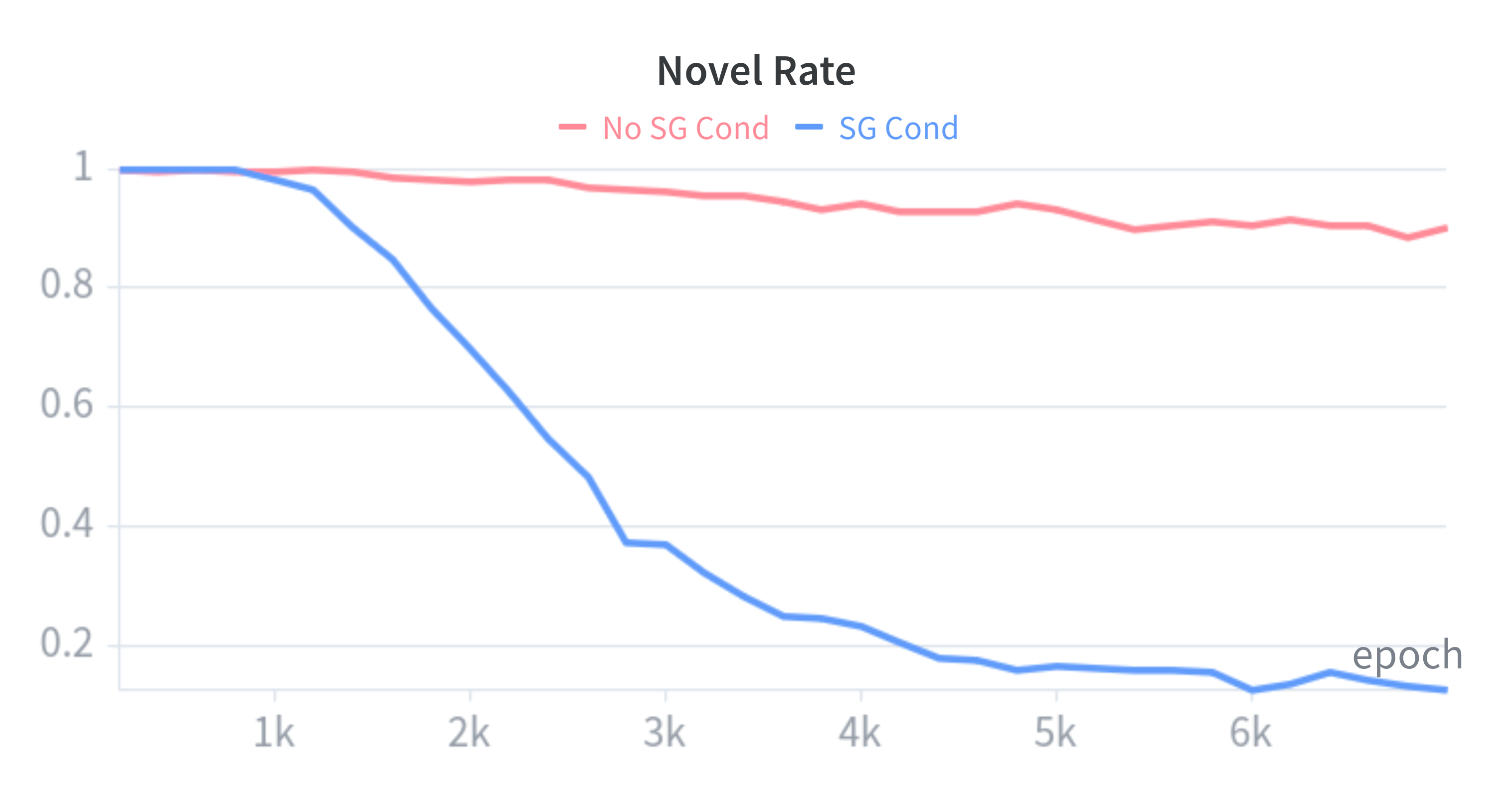}
        \caption{Novelty}
        \label{fig:with_sg}
    \end{subfigure}
    \caption{Comparison of structural validity and novelty for DiT models with and without space group conditioning.}
    \label{fig:ablation_sg_cond}
\end{figure}

In this section, we examine the role of space group conditioning in our framework. We modify Algorithm \ref{alg:ae_encoder} by removing the space group encoding to train a new autoencoder. We then train two latent flow matching models: one with space group conditioning and one without. The results are shown in Figure \ref{fig:ablation_sg_cond}. Without conditioning, the model achieves high novelty but low structural validity. In contrast, incorporating space group conditioning significantly improves structural validity, albeit at the cost of reduced novelty. This highlights the importance of space group conditioning in our representation, in contrast to ADiT, which uses the same conditioning but is only able to generate mostly one space group (P1). Furthermore, this also suggests an interesting avenue for future research: identifying additional sources of information that may be beneficial to condition on.

\section{DFT Details}
\label{app:dft_details}
First-principles structural optimisations were conducted in the framework of density functional theory (DFT) \cite{kohn1965self,hohenberg1964inhomogeneous} using the Vienna \textit{Ab initio} Simulation Package (VASP) \cite{kresse1993ab,kresse1994ab,kresse1996efficiency,kresse1996efficient}. The exchange-correlation potential was treated using the generalized gradient approximation (GGA) functional in conjunction with the Perdew, Burke, and Ernzerhof (PBE) method \cite{perdew1996generalized}. The projector augmented wave method \cite{Blochl1994projector} was applied. All input parameters, including plane-wave cutoffs, convergence criteria, and k-point densities, were selected using pymatgen’s MPRelaxSet \cite{pymatgen}. For systems where the tetrahedron smearing method led to convergence issues, Gaussian smearing with a smearing width of 0.05 was employed. Thermodynamic stability was assessed by constructing the convex hull relative to all competing phases reported in the Materials Project MP20 and MPTS52 datasets. To enable direct comparison with Materials Project data, the MP energy correction scheme was employed for anions as well as for combining GGA and GGA+$U$ calculations \cite{jain2011high, jain2011formation}. The convex hull analysis was performed using the pymatgen library \cite{pymatgen}. The convex hull was constructed without incorporating the energies of the newly predicted structures. Accordingly, phases lying on or below the hull are considered thermodynamically stable, while those with energies up to 0.1 eV/atom above the hull are classified as metastable. The reported energy difference with respect to the convex hull is equivalent to the formation enthalpy ($\Delta H_{cp}$) defined relative to the set of competing phases, as commonly used in the materials science literature~\cite{dahlqvist2024max}.

\section{Evaluation Metrics}
\label{app:eval_metrics}
Similar to prior works ADiT \cite{adit}, we follow the evaluation protocol established by \cite{flow_matching}, where we sample $10,000$ crystals and compute validity, stability, uniqueness, and novelty, as defined as follows:

\begin{itemize}
    \item Structural Validity: \% of crystals with all pairwise distances >= 0.5 and crystal volume >= 0.1.

    \item Compositional Validity: \% of crystal compositions with charge neutrality and electronegativity balance according to SMACT \cite{smact}. As noted in the main paper, only approximately 90\% of the MP20 crystals pass this check; therefore, it should not be relied upon heavily.

    \item Uniqueness: \% of generated crystals that are not duplicates of each other. Comparisons were performed using StructureMatcher from pymatgen \cite{pymatgen}. 

    \item Novelty: \% of generated crystals that are not duplicates of the training data. Comparisons were performed using StructureMatcher from pymatgen \cite{pymatgen}.

    \item Stability: \% of crystals with $\Delta H_{cp}$ <0.0 eV/atom and number of unique elements >= 2. (We also report metastability as $\Delta H_{cp}$ <0.1 eV/atom and number of unique elements >= 2.)

    \item U.N. Rate: \% of structurally valid crystals that are both unique and novel

    \item \sun: \# of structurally valid crystals that are stable (See Appendix \ref{app:dft_details}), unique and novel. 

    \item \msun: \# of crystals that are metastable (See Appendix \ref{app:dft_details}), unique and novel.

    \item P1 rate: the fraction of the 10,000 generated crystals classified as P1, as determined by pymatgen \cite{pymatgen}.   
\end{itemize}

\section{Licenses}
\label{app:licenses}
\paragraph{Code} We have used the ADiT codebase \footnote{\url{https://github.com/facebookresearch/all-atom-diffusion-transformer}} as starting point for our implementation. This code is available under an MIT license. 

\paragraph{Training data} We used datasets derived from Materials Project \cite{materials_project1,materials_project2}, which provide their data under a CC-BY 4.0 license. \footnote{\url{https://creativecommons.org/licenses/by/4.0/}}. 

\paragraph{Data from compared methods} Generated structures for compared methods, except ADiT, SymmCD, SGEquiDiff, and OMatG were provided by \citet{crystal_dit}, available online\footnote{\url{https://huggingface.co/xiaohan-yi/CrystalDiT}} under a MIT license. For ADiT, SymmCD, SGEquiDiff, and OMatG, we trained models from scratch to generate structures ourselves using official public implementations, without any changes.

\section{Space group Distribution of Generated Structures}
\label{app:gen_sg_dist}

\begin{figure}[H]
    \centering
    \includegraphics[width=0.9\linewidth]{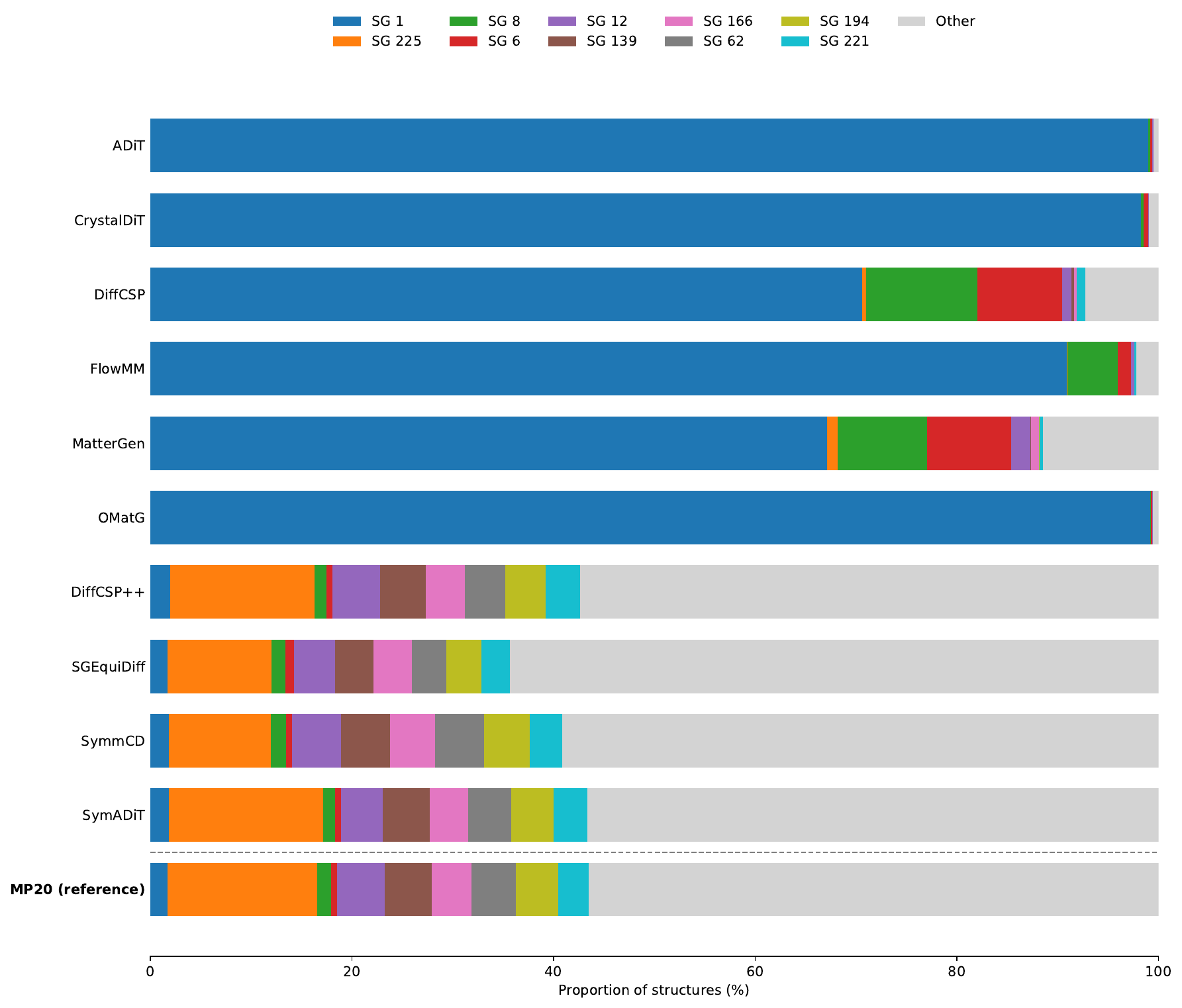}
    \caption{
        Space group distributions of structures generated by each method in relation to the MP20 dataset.
        Each horizontal bar represents the full set of generated structures for a given method, with the width of each colored segment indicating the fraction assigned to the corresponding space group.
        Colors denote the ten most frequent space groups across all evaluated methods, while all remaining space groups are grouped into \textit{Other}.
        }
    \label{fig:spacegroup_distribution}
\end{figure}

To showcase clearly the diversity of generated crystal symmetries, and further support the results in Table \ref{tab:mp20_results}, we compare the space group distributions produced by each method to the MP20 dataset. Results can be seen in Figure ~\ref{fig:spacegroup_distribution}. The figure shows the fraction of generated structures assigned to each of the ten most frequent space groups across the evaluated methods, with all remaining space groups grouped into \textit{Other}. We note that SymADiT best follows the distribution of the reference MP20 dataset. This suggests that, for some competing symmetry-aware methods, the final generated structures may not always preserve the space group initially sampled during generation, thereby altering the resulting space-group distribution.

\section{Novel Crystal Statistics}
\label{app:additional_results}

In this section, we present some statistical analyses of the novel structures generated by each model, based on 1,000 samples drawn from the set of structurally valid and unique crystals, and compare them to the MP20 training set.

Table \ref{tab:unique_elements_stats} showcases compositions in terms of the number of unique elements per structure. Overall, all methods exhibit similar statistics, with mean values close to the training distribution, indicating that compositional complexity is largely preserved across models. SymADiT matches best with the training distribution.

Table \ref{tab:rare_earth_stats} reports the occurrence of rare-earth elements across generated structures. RE denotes the number of generated structures that contain at least one rare-earth element, while \%RE indicates the corresponding percentage relative to the total number of structures. Mean RE measures the average number of distinct rare-earth elements per structure, providing a finer-grained view of how heavily such elements are incorporated. We observe that most methods produce comparable percentages to the training data, suggesting that rare-earth inclusion is largely consistent across approaches. Notably, CrystalDiT exhibits a higher proportion and mean count of rare-earth elements, which may explain its increased novelty and could also contribute to its high M/S.U.N. metrics, since fewer comparable crystals are available in the convex hull calculations.

Table \ref{tab:num_orbits_stats} reveals a pronounced difference in structural complexity across methods. Symmetry-agnostic models generate structures with substantially higher orbit counts, while symmetry-aware methods produce distributions that closely align with the training data, suggesting more physically plausible crystal structures. This can be explained by the fact that in P1 structures, which symmetry-agnostic methods predominantly sample from, each atom forms its own symmetry orbit, leading to inflated orbit counts.

\begin{table*}[h]
\centering
\footnotesize

\begin{minipage}[H]{0.48\textwidth}
\centering

\centering
\scriptsize
\caption{\textbf{Composition complexity.} Statistics of the number of unique element types per generated structure.}
\label{tab:unique_elements_stats}
\setlength{\tabcolsep}{3.0pt}
\renewcommand{\arraystretch}{1.25}

\begin{tabular}{l c c c c c c}
\toprule
Method & Sym. & \#Struct. & Mean & Std. & Min & Max \\
\midrule
SymmCD      & Yes & 927  & 3.522 & 1.241 & 1 & 9 \\
DiffCSP++   & Yes & 898  & 3.477 & 1.278 & 1 & 10 \\
CrystalDiT  & No  & 651  & 3.353 & 1.016 & 1 & 7 \\
DiffCSP     & No  & 911  & 3.334 & 1.066 & 1 & 8 \\
MatterGen   & No  & 918  & 3.215 & 0.874 & 1 & 7 \\
SGEquiDiff  & Yes & 895  & 3.179 & 0.938 & 1 & 8 \\
FlowMM      & No  & 909  & 3.164 & 0.801 & 1 & 6 \\
ADiT        & No  & 486  & 3.156 & 0.765 & 1 & 5 \\
\textbf{SymADiT} & Yes & 546  & 3.147 & 0.764 & 1 & 6 \\
\midrule
MP20 (train) & Yes & 27138 & 3.014 & 0.726 & 1 & 7 \\
\bottomrule
\end{tabular}

\end{minipage}
\hfill
\begin{minipage}[H]{0.48\textwidth}
\centering

\centering
\scriptsize
\caption{\textbf{Rare-earth composition.} Statistics of rare-earth element occurrence in generated structures. }
\label{tab:rare_earth_stats}
\setlength{\tabcolsep}{3.0pt}
\renewcommand{\arraystretch}{1.25}

\begin{tabular}{l c c c c c}
\toprule
Method & Sym. & \#Struct. & \#RE & \%RE & Mean RE \\
\midrule
SGEquiDiff  & Yes & 895  & 307 & 34.30 & 0.406 \\
SymmCD      & Yes & 927  & 301 & 32.47 & 0.392 \\
DiffCSP++   & Yes & 898  & 374 & 41.65 & 0.490 \\
\textbf{SymADiT} & Yes & 546  & 215 & 39.38 & 0.430 \\
ADiT        & No  & 486  & 192 & 39.51 & 0.422 \\
CrystalDiT  & No  & 651  & 310 & 47.62 & 0.922 \\
MatterGen   & No  & 918  & 362 & 39.43 & 0.460 \\
FlowMM      & No  & 909  & 390 & 42.90 & 0.470 \\
DiffCSP     & No  & 911  & 288 & 31.61 & 0.384 \\
\midrule
MP20 (train) & Yes & 27138 & 11018 & 40.60 & 0.440 \\
\bottomrule
\end{tabular}

\end{minipage}

\vspace{0.5cm}

\begin{minipage}[H]{0.6\textwidth}
\centering

\centering
\scriptsize
\caption{\textbf{Structural complexity.} Statistics of the number of symmetry orbits per generated structure.}
\label{tab:num_orbits_stats}
\setlength{\tabcolsep}{3.0pt}
\renewcommand{\arraystretch}{1.25}

\begin{tabular}{l c c c c c c}
\toprule
Method & Sym. & \#Struct. & Mean & Std. & Min & Max \\
\midrule
ADiT        & No  & 486  & 12.366 & 5.074 & 1 & 20 \\
CrystalDiT  & No  & 651  & 10.931 & 5.675 & 1 & 20 \\
DiffCSP     & No  & 911  & 10.692 & 5.593 & 1 & 20 \\
FlowMM      & No  & 909  & 10.411 & 5.327 & 1 & 20 \\
MatterGen   & No  & 918  & 9.584  & 4.925 & 1 & 20 \\
\textbf{SymADiT} & Yes & 546  & 5.328 & 3.260 & 1 & 20 \\
SymmCD      & Yes & 927  & 5.028 & 3.284 & 1 & 20 \\
DiffCSP++   & Yes & 898  & 5.027 & 3.203 & 1 & 20 \\
SGEquiDiff  & Yes & 895  & 4.667 & 2.761 & 1 & 20 \\
\midrule
MP20 (train) & Yes & 27138 & 4.758 & 3.012 & 1 & 20 \\
\bottomrule
\end{tabular}

\end{minipage}

\label{tab:appendix_stats}
\end{table*}

\section{Statistical Analysis}
\label{app:statistical_analysis}

Table~\ref{tab:stats} summarizes the outcomes of the 500-sample evaluation and reports the corresponding Wilson score confidence intervals. SymADiT achieves the highest number of \morsun crystals among the symmetry-aware approaches, showing its strong empirical performance so far in the crystal generation pipeline. 

The \(95\%\) Wilson confidence intervals for the \morsun rates of SymADiT overlap with those of most competing methods. As a result, the observed differences cannot generally be considered statistically distinguishable, and the confidence-interval analysis alone does not provide sufficient evidence that SymADiT consistently outperforms or underperforms the evaluated models. 

To further assess statistical significance, we conducted pairwise two-proportion \(z\)-tests on the S.U.N. outcomes, using SymADiT as the reference method. The improvements over SymmCD, DiffCSP, CrystalDiT, MatterGen, and FlowMM are statistically significant (\(p < 0.01\)). The improvement over DiffCSP++ is also notable, although it falls short of the conventional significance threshold (\(p = 0.061\)). In contrast, no statistically significant difference is detected between SymADiT and its non-symmetry-aware counterpart, ADiT (\(p = 0.145\)).

\begin{table*}[h]
\centering
\footnotesize
\caption{
Number of \sun and \msun crystals, with Wilson 95\% confidence intervals (CI).
The $z$- and $p$-values are obtained from pairwise two-proportion tests on \sun using SymADiT as the reference.}
\label{tab:stats}

\renewcommand{\arraystretch}{1.2}

\begin{tabular*}{\textwidth}{@{\extracolsep{\fill}} l cc cc cc}
\toprule
Method
& \# \sun $\uparrow$
& \# \msun $\uparrow$
& \sun W-CI
& \msun W-CI
& $z$
& $p$ \\
\midrule
SGEquiDiff
& 51
& 89
& 7.8--13.2\%
& 14.7--21.4\%
& $-3.14$
& 0.002 \\

SymmCD
& 49
& 78
& 7.5--12.7\%
& 12.7--19.0\%
& $-3.34$
& $<0.001$ \\

DiffCSP++
& 64
& 105
& 10.2--16.0\%
& 17.7--24.8\%
& $-1.87$
& 0.061 \\

ADiT
& 103
& 170
& 17.3--24.4\%
& 30.0--38.3\%
& $+1.46$
& 0.145 \\

CrystalDiT
& 47
& 152
& 7.1--12.3\%
& 26.5--34.6\%
& $-3.55$
& $<0.001$ \\

MatterGen
& 41
& 126
& 6.1--10.9\%
& 21.6--29.2\%
& $-4.19$
& $<0.001$ \\

FlowMM
& 36
& 106
& 5.2--9.8\%
& 17.8--25.0\%
& $-4.75$
& $<0.001$ \\

DiffCSP
& 53
& 107
& 8.2--13.6\%
& 18.0--25.2\%
& $-2.93$
& 0.003 \\

OMatG
& 87
& 134
& 14.3--21.0\%
& 23.1--30.8\%
& $+0.17$
& 0.867 \\

SymADiT
& 85
& 136
& 14.0--20.5\%
& 23.5--31.3\%
& 0.00
& -- \\

\bottomrule
\end{tabular*}
\end{table*}

\section{A Discussion regarding the Novelty Metric}
\label{app:novelty_discussion}

In particular, generating a structure that is classified as novel can, in principle, be relatively straightforward: sufficiently perturbing the atomic configuration of a known structure may cause it to no longer match any structure in the training set. Whether such a structure is physically realistic, however, is a separate question, for which stability provides an important indication. This distinction is particularly relevant for the \morsun metric, which is also commonly reported in the literature. Typically, \morsun is reported as a conditional fraction among novel structures. Consequently, a high novelty rate can increase the pool of structures entering this metric without necessarily implying that these structures are physically plausible. Such candidates may ultimately fail the subsequent DFT-based stability evaluation.

For this reason, we believe it is more important to report the absolute number of evaluated structures that ultimately satisfy the stability criteria, rather than a conditional fraction. A high novelty rate alone can otherwise give an overly favourable impression if many of the novel structures are physically implausible and subsequently rejected by DFT.

This does not diminish the value of novelty as an evaluation metric. In particular, novelty is useful when combined with computationally inexpensive screening methods, where it can serve as an effective precursor to more expensive DFT calculations by identifying candidates that are distinct from the training data.

Our point is therefore not that novelty should be disregarded, but rather that it should be \emph{interpreted jointly with stability and with the number of structures that successfully pass the full evaluation pipeline}. A high novelty rate by itself does not necessarily imply that a model generates a larger number of physically plausible new crystals.

\section{Comparison of Inference Speed between SymADiT and ADiT}
\label{app:inference_speed}
Comparison of inference speeds between models is not a trivial task -- as it is unclear what constitutes a well defined measurement (e.g., time to a single forward pass vs time to a fully generated CIF file). Furthermore, each method applies different post-processing steps. The exception in this work is the comparison between SymADiT and ADiT. The two generation pipelines are similar in how they simulate the ODE and apply the decoder, enabling a meaningful comparison. This experiment will also demonstrate the inference-speed benefits of reducing the number of tokens, specifically by using symmetry orbits as Transformer tokens rather than individual atoms.

For both models, we generate 10,000 crystals on a single NVIDIA H200 GPU using a maximum batch size of 1,024 and 500 ODE steps. The models also have comparable parameter counts. To make the timing comparison as consistent as possible, we exclude model-specific sanity checks and filtering, such as implausible lattice parameters or atomic overlaps, since these differ between the pipelines. For ADiT, we evaluate only periodic crystal generation and exclude molecular generation.

We decided to divide the evaluation of inference time into two parts:
    \begin{enumerate}
        \item Model inference. We measure the time required to produce the 10,000 decoded outputs needed to construct all crystals. This includes the iterative ODE sampling procedure and the VAE decoder at the end.
    
        \item Post-processing. We measure the time required to convert the 10,000 decoded model outputs into explicit pymatgen crystal structures. Here, the two models require different post-processing steps because their outputs represent different quantities. For example, SymADiT predicts Wyckoff positions and needs to post-process through Pyxtal, whereas ADiT does not.
    \end{enumerate}

\begin{table*}[h]
\centering
\footnotesize
\caption{\textbf{Generation time.} Model inference, post-processing, and total generation time for \thsnd{10} generated crystals. All times are reported in seconds.}
\label{tab:inference_speed}
\setlength{\tabcolsep}{6.0pt}
\renewcommand{\arraystretch}{1.30}

\begin{tabular}{l ccc}
\toprule
Method & Model inference (s) & Post-processing (s) & Total time (s) \\
\midrule
SymADiT
& 263.9
& 90.8
& 354.7 \\

ADiT
& 849.4
& 2.3
& 851.7 \\
\bottomrule
\end{tabular}
\end{table*}

Results are presented in Table \ref{tab:inference_speed}. For model inference, SymADiT achieves 3.22x speedup in inference time. The improvement is consistent with SymADiT representing crystals using symmetry orbits rather than individual atoms, which on average results in fewer tokens and therefore reduces the computational cost of the iterative denoising process.

For post-processing, ADiT is much faster, as it can go straight-forwardly from the model output to a pymatgen crystal structure, while in our implementation we rely on converting the model output from ASU to full unit cell via PyXtal, which is slower.

Despite the longer post-processing time, SymADiT still achieves a 2.40× inference speed-up over ADiT.

\section{Visual Results}
\label{app:visual_results}
\begin{figure}[H]
    \centering
    \includegraphics[width=\textwidth]{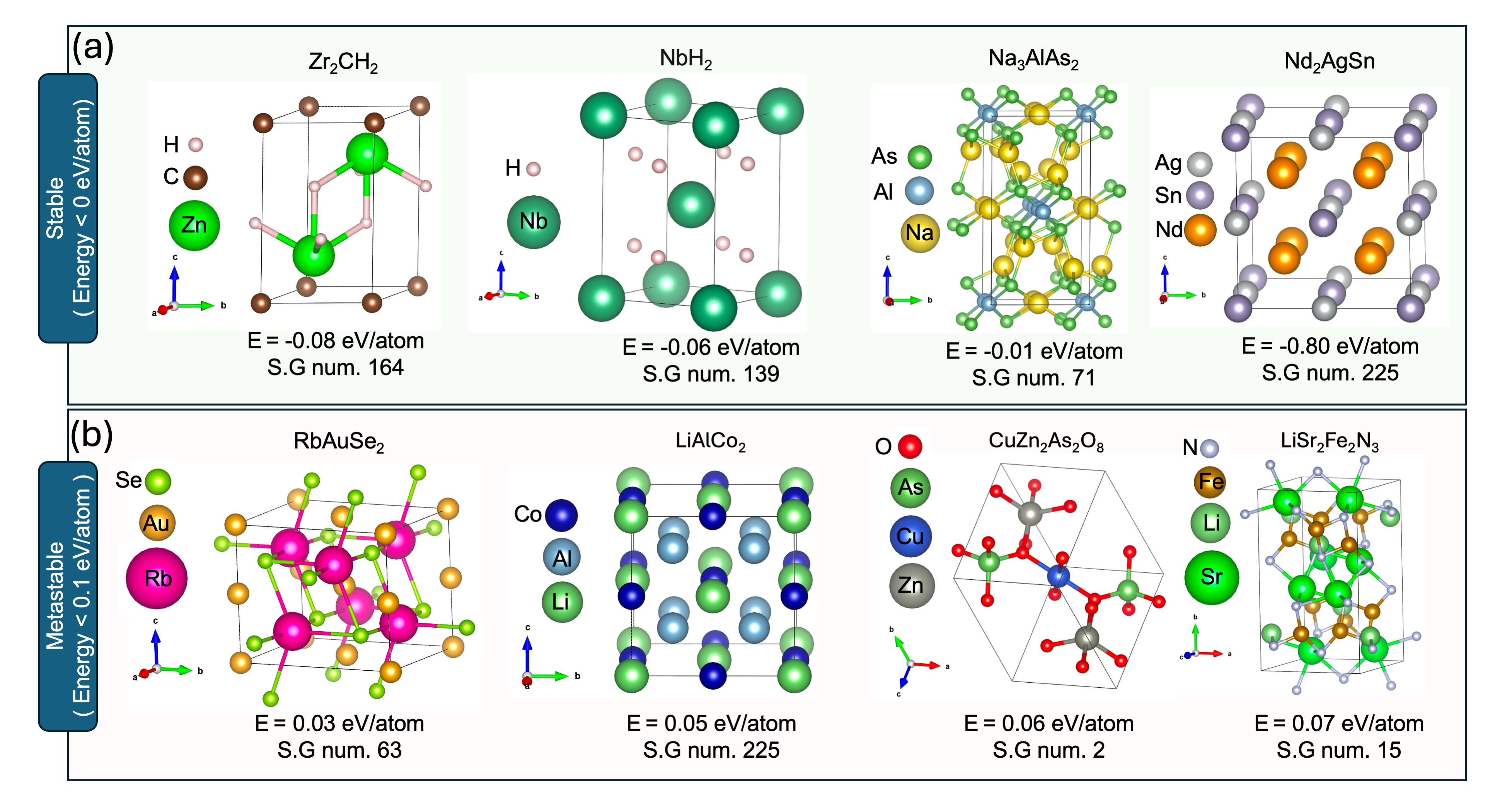}
    \caption{Schematic illustration of crystals identified by SymADiT: (a) thermodynamically stable structures (S.U.N) with $\Delta H_{cp} < 0$, and (b) metastable structures \morsun with $0 \leq \Delta H_{cp} < 0.1$ eV/atom.}
    \label{fig:visual_results}
\end{figure}

In Figure \ref{fig:visual_results}, we present S.U.N and M.S.U.N crystals from SymADiT. The generated structures show a wide range of space groups and compositions. Visualisation was done using the VESTA software \cite{vesta}.

\end{document}